\documentclass[APA,LATO1COL]{WileyNJD-v2}

\usepackage{amsmath,amssymb}
\usepackage{color,xcolor}
\usepackage{enumerate}
\usepackage{setspace}
\usepackage{lastpage}
\usepackage{graphicx}
\usepackage{adjustbox}
\usepackage{float}
\usepackage{xspace}
\usepackage{booktabs}
\usepackage{url}
\usepackage{numprint}
\npthousandsep{,}
\usepackage{algpseudocode}
\usepackage{algorithm}

\usepackage{calc}
\usepackage{xstring}

\makeatletter
\newcommand*{\toccontents}{\@starttoc{toc}}
\makeatother

\DeclareGraphicsExtensions{.pdf,.png,.eps}

\newcommand{\eg}{e.\,g.}
\newcommand{\ie}{i.\,e.}

\let\originalleft\left
\let\originalright\right
\renewcommand{\left}{\mathopen{}\mathclose\bgroup\originalleft}
\renewcommand{\right}{\aftergroup\egroup\originalright}

\makeatletter

\let\oldnorm\|

\definecolor{orange}{rgb}{1,.647,0}

\renewcommand{\|}{|\!|}         
\newcommand{\T}{{}^{\mathsf{T}}}

\DeclareMathOperator*{\argmax}{argmax}

\DeclareMathOperator{\trace}{trace}                  
\DeclareMathOperator{\EDF}{EDF}

\DeclareMathOperator{\mean}{mean}
\DeclareMathOperator{\sd}{sd}

\newcommand{\cF}{{\cal{F}}}

\newcommand{\cO}{{\cal{O}}}

\newcommand{\IR}{{\mathbb{R}}}

\newcommand{\bxi}{{\boldsymbol{\xi}}}

\newcommand{\bSigma}{{\boldsymbol{\Sigma}}{}}

\newcommand{\I}{{\mathbf{I}}}

\renewcommand{\P}{{\mathbf{P}}}  

\renewcommand{\S}{{\mathbf{S}}}

\newcommand{\0}{{\mathbf{0}}}

\newcommand{\w}{{\textbf{\textit{w}}}}
\newcommand{\x}{{\textbf{\textit{x}}}}
\newcommand{\y}{{\textbf{\textit{y}}}}

\newcommand*{\stack@relbin}[3][]{%
  \mathop{#3}\limits
  \toks@{#1}%
  \edef\reserved@a{\the\toks@}%
  \ifx\reserved@a\@empty\else_{#1}\fi
  \toks@{#2}%
  \edef\reserved@a{\the\toks@}%
  \ifx\reserved@a\@empty\else^{#2}\fi
  \egroup
}%
\renewcommand*{\stackrel}{%
  \mathrel\bgroup\stack@relbin
}
\newcommand*{\stackbin}{%
  \mathbin\bgroup\stack@relbin
}

\articletype{Article Type}%

\received{26 April 2016}
\revised{6 June 2016}
\accepted{6 June 2016}

\begin{document}

\title{Fast covariance parameter estimation of spatial Gaussian process models using neural networks}

\author[1]{Florian Gerber*}

\author[1]{Douglas W.\ Nychka}

\authormark{F.\ Gerber and D.\ W.\ Nychka}

\address[]{\orgdiv{Department of Applied Mathematics and Statistics}, \orgname{Colorado School of Mines}, \orgaddress{1500 Illinois St., Golden \state{CO} 80401, \country{USA}}}

\corres{*\email{gerber@mines.edu}}


\abstract[Summary]{
  Gaussian processes (GPs) are a popular model for spatially referenced data and allow descriptive statements, predictions at new locations, and simulation of new fields.
Often a few parameters are sufficient to parameterize the covariance function, and maximum likelihood (ML) methods can be used to estimate these parameters from data.
ML methods, however, are computationally demanding. 
For example, in the case of local likelihood estimation, even fitting covariance models on modest size windows can overwhelm typical computational resources for data analysis.  
This limitation motivates the idea of using neural network (NN) methods to approximate ML estimates. 
We train NNs to take moderate size spatial fields or variograms as input and return the range and noise-to-signal covariance parameters. 
Once trained, the NNs provide estimates with a similar accuracy compared to ML estimation and at a speedup by a factor of 100 or more.
Although we focus on a specific covariance estimation problem motivated by a climate science application, this work can be easily extended to other, more complex, spatial problems and provides a proof-of-concept for this use of machine learning in computational statistics.
}

\keywords{Climate model, Gaussian process, neural network, non-stationary covariance, spatial statistics, TensorFlow}


\maketitle

\footnotetext{\textbf{Abbreviations:} EDF, effective degrees of freedom; GP, Gaussian process; ML: maximum likelihood; NN, neural network; }

\vspace*{1cm}
\section{Introduction}\label{sec:1}

A benefit of the rapid advances in computing, data storage, and sensor technology is the availability of large datasets to address substantial scientific questions. 
Such data are often spatially referenced and relevant in weather, climate, remote sensing applications, and other areas.
A key characteristic of many of these datasets is the large number of observations, which poses statistical and computational challenges in their analysis.
This impediment is notably present for the popular GP framework, which parameterizes the covariance structure with a few parameters and then estimates them, \eg, by~ML.
Using exact computations, one evaluation of that likelihood function requires $\cO\left(n^3\right)$ operations and $\cO\left(n^2\right)$ of memory, where $n$ is the number of observations.
Moreover, the ML optimization problem often does not have a closed-form solution, and hence, numerical optimizers are used and so require \emph{many} evaluations of the loss function.
This computational burden hinders the covariance parameter estimation and exploratory analysis for large and heterogeneous spatial data. 
Although there is a large amount of research on GP models aiming at a fast approximate inference,
the available methods are still computationally expensive and often require a dedicated (high-performance) computing infrastructure \citep{Heat:Gerb:etal:18,Liu:etal:20,gerb:nych:20}. In addition, given large and non-stationary spatial data global covariance models may not be as useful as fitting non-stationary covariance functions based on local tiles or windows of the data. 
Lack of efficient approximate statistical approaches may explain why some applied scientists turn away from GP models in favor of algorithmic methods designed for specific tasks.
For example, there are many recent algorithms for the fast prediction of missing observations in satellite data; see Table~1 in \cite{Gerb:Jong:Scha:Scha:Furr:18} for an overview.
Such methods are typically much faster but provide only a limited statistical framework, which makes uncertainty quantification difficult.

We view remotely sensed and physically derived fields (images) as a specific type of spatial data, and recent advances in NN methodology enable the efficient processing of large data volumes thereof \citep{Schm:15,Ojha:etal:17,Yuan:etal:20}.
Especially for categorizing images, convolutional NNs have been successful \citep{Kriz:etal:17}.
The NN literature reveals, however, that these flexible, functional representations can also be used for covariance structures of time-series data \citep{Pete:96,crem:etal:17,Liu:etal:18} and for kernel estimation in the field of numerics \citep{dask:20}. 
However, to our knowledge, NNs have not been used to estimate covariance parameters of GP models for spatial data. 
This gap has motivated the research in this article.

In most applications, NNs are used as an algorithm for prediction independent of a statistical framework.
However, in this work, we take a different perspective, where the target statistical computation is well defined and the NNs are used to provide a fast and accurate approximation. 
To explain this further, consider ML estimation as a function taking data as input and returning the covariance parameters.
The ML estimates are derived from assumptions on the data model, and their computation for covariance function parameters involves a possibly complicated optimization.
But the function is fixed, \ie, it has no free parameters. 
Similarly, the NN approach sees the estimation as the evaluation of a function taking data as input and returning the covariance parameters.
However, with this approach, the function is \emph{not} derived from assumptions on the data generation process.
Rather it is designed to have a computationally appealing form and weights (parameters) allowing it to represent a large class of functions.
The weights are trained (estimated) from training samples consisting of synthetically generated fields and their corresponding, known covariance parameters.
This way, the NN learns to represent covariance parameter estimation, and the accuracy of its estimates can be tested against another sample of synthetically generated fields.
Moreover, the accuracy of the NN can be assessed against the exact ML estimates, which we assume to be unique and close to an optimal choice in the absence of prior information. 
At the outset, it may be unclear why NNs should perform well at determining covariance parameters from spatial fields, and the ML estimates set a high bar for accuracy.
Despite this initial doubt, we found the NN to be amazingly successful at this task.

But why go to the trouble of training NNs just to reproduce ML estimates?
In this work, we show that a NN can be evaluated much more rapidly than direct ML optimization.
Once the NN is trained, this approximation can give a factor of 100 or more speedup in determining covariance parameters of a typical GP covariance function.
Our experience is that when a statistical analysis enjoys this kind of increase in speed, it opens up novel ways of exploratory analyses, resampling, and modeling.
In particular, having a computational tool for rapid local likelihood fitting will offer new ways to consider non-stationary processes. 

This study was motivated by local covariance estimation in climate science, which is used to emulate large, Gaussian fields generated from climate model experiments. 
In a nutshell, the context of this data application is as follows:
Climate model simulations are often computationally expensive, running on dedicated high-performance computers, and their size limits the number of feasible model runs.
At the same time, a suite of climate model runs with perturbed initial conditions are important to reveal the internal variability of the climate system.
The NCAR Large Ensemble Project (NCAR-LENS) \citep{Kay:etal:15} studies this variability and provides the model output from $30+$ model runs with perturbed conditions. Our particular case study from NCAR-LENS considers 
$30$ {\it pattern scaling }fields, which describe the expected local change of the mean surface temperature for June, July, and August resulting from a one-degree increase of the global mean temperature.
\cite{Nych:etal:18} then use a GP model to emulate similar fields at relatively low computational cost \citep{Alex:18}.
A key component of their model is the description of a non-stationarity covariance function of the fields derived from a moving window, maximum likelihood estimator. 
More precisely, they estimate the covariance parameters of a GP model at each model grid point based on an $11 \times 11$ square window neighborhood and using the 30 replicate fields. 
This is a computationally intensive task, and \cite{Nych:etal:18} rely on high-performance computing with up to $1000$ CPUs to accomplish it.
We show that using NN models similar parameter estimates can be obtained on a laptop within minutes.

The paper is structured as follows:
Section~\ref{sec:2} introduces different NN and ML-based covariance parameter estimation methods.
Section~\ref{sec:3} compares the methods using simulated data.
Section~\ref{sec:4} compares one NN and one ML-based method using the mentioned estimation task relevant to climate science.
Finally, Section~\ref{sec:5} discusses the advantages, challenges, and future research opportunities of NN based covariance parameter estimation.

\section{Method}\label{sec:2}
We consider spatial data on a regularly spaced grid of locations and use a GP to describe the spatial dependence structure.
Although our method can generalize to irregularly spaced locations, a regular grid is used for a proof-of-concept and is adequate for the climate model example.
Assume the spatial field consists of $n$ values $\y =(y_{1}, \dots, y_{n})\T\in\IR^n$ at spatial locations $\{ s_1, \dots, s_{n} \} $. 
We assume $\y$ to be distributed~as
\begin{equation}\label{eq:mod}
  \begin{split}
    \y \sim \mathcal{N}\left(\0, \sigma^2 \bSigma(\theta) + \tau^2 \I  \right),
  \end{split}
\end{equation}%
where $\sigma^2>0$ is the marginal variance (partial sill), $\tau^2 \geq 0$ is the measurement error (nugget effect), and $\theta$ are parameters of the $n \times n$ covariance matrix~$\bSigma(\theta)$ derived from a covariance function $c(s, s^\prime, \theta)$ and so explicitly $\bSigma(\theta)_{i,j} = c(s_i, s_j, \theta)$.
In this work, $\theta$ is a scalar range parameter. 

\subsection{ML framework and outline of the estimation task}\label{sec:ml}
Based on \eqref{eq:mod}, the log-likelihood function for the parameters $\bxi=(\sigma^2, \tau^2, \theta)\T$ and spatial field $\y$~is 
\begin{equation}
\label{eq:lik}
    l(\bxi; \y)=  -\frac12 \left[\y^\top\left(\sigma^2\bSigma(\theta) + \tau^2\I\right)^{-1} \y+ \log \det\left(  \sigma^2\bSigma(\theta)+ \tau^2\I \right) +n\log(2\pi)\right] ,\\
\end{equation}  
and the corresponding ML estimate $\widehat \bxi$ is the global maximizer of~\eqref{eq:lik} with respect to~$\bxi$.
In other words, the ML estimation is a function taking $\y$ as input and mapping it to~$\widehat \bxi$ by
\begin{equation}\label{eq:layer}
  {\cal F} _\text{ML}: \IR^n \rightarrow D,\quad {\cal F}_\text{ML}(\y) = \argmax_{\bxi} \,l(\bxi; \y) = \widehat \bxi,
\end{equation}
where $D$ denotes the valid parameter space of~$\bxi$.
As cited in the introduction, the computational cost of this maximization rapidly increases as $n$ becomes large. We note, however, that based on practical experience, the log-likelihood surface for covariance parameters tends to be smooth and with a finite single maximum or a maximum in the limit at infinity. 
This motivates the search for a computationally more efficient function mapping~$\y$ to~$\widehat \bxi$, and we use the NN framework to construct~it. 

An important simplification for this standard problem, and how we report our results, is to reparametrize the likelihood in terms of $\lambda= \tau^2/ \sigma^2 $,~$\theta$, and $\sigma^2$. Recall that the ML estimates are invariant under 1-1 transformations of the parameters, and hence, this does not change the estimates of~$\xi$.
The benefit of this reparametrization is that the log-likelihood can be maximized analytically for $\sigma^2$ given the other two parameters, \ie, 
\begin{equation}
\widehat{ \sigma}^2( \lambda, \theta) = \frac{ \y^\top\left( \bSigma(\theta) + \lambda \I\right)^{-1} \y} {n}.
\end{equation}
Substituting $\widehat{\sigma}^2( \lambda, \theta)$ back into the log-likelihood yields a profile likelihood, which is concentrated on just $\lambda$ and~$\theta$. 
Finally, for independent replicates of fields from the same data generation process, the joint log-likelihood is obtained by summing the individual log-likelihoods.
The concentration to $\lambda$ and $\theta$ also goes through in that case, and $\widehat{\sigma}^2( \lambda, \theta)$ becomes the mean of the individual estimates. 
We denote this model by \emph{ML} if the input sample consists of one field $\y$ and by \emph{ML30} if the input sample consists of $30$ replicated fields $\{\y_1, \dots, \y_{30}\}$.
We will match both models with two NN models designed to accomplish the same estimation task; see Table~\ref{tab:models} for an overview.

\begin{table}[b]
 \caption{The six models used to estimate the covariance parameters $\bxi=(\lambda, \theta)\T$.}
 \label{tab:models}
   \centering
    \begin{tabular}{lll}
        \toprule
name             &        description                                                         & input              \\    
\midrule
ML               &   ML estimation          & one $16\times 16$ field               \\
ML30             &   ML estimation          & $30$ $16\times 16$ fields               \\
NF               &   convolutional NN based on fields       & one $16\times 16$ field              \\
NF30             &   convolutional NN based on fields       & $30$ $16\times 16$ fields               \\
NV               &   dense NN based on variograms & one $16\times 16$ field               \\
NV30             &   dense NN based on variograms & $30$ $16\times 16$ fields               \\
\bottomrule
    \end{tabular}
\end{table}

The connection between data smoothing and spatial statistics suggest a further parameterization, which aids in interpreting~$\lambda$.
In the presence of measurement error, \ie, $\tau^2 > 0$, the predicted curve can be associated with an effective degrees of freedom (EDF) given by 
\begin{equation}
  \EDF(\lambda, \theta) =  \trace \left[  \bSigma(\theta) \left( \bSigma(\theta) + \lambda \I \right)^{-1} \right],
\end{equation}
and by linear algebra, it can be shown that $\EDF(\lambda, \theta) \in [0, n]$. 
For fixed~$\theta$, $\EDF(\lambda, \theta)$ is 1-1 in $\lambda$, and thus, this gives yet another reparametrization in $\EDF$ and~$\theta$.
That is, given $\EDF$ and $\theta$ one can identify a corresponding pair of $\lambda$ and~$\theta$.
We use this version to create representative grids of the $\lambda$ parameter to estimate parameters via grid search and to train the NNs. 

\begin{figure}
    \centering
    \hspace*{2cm}
    \includegraphics[width=.6\textwidth]{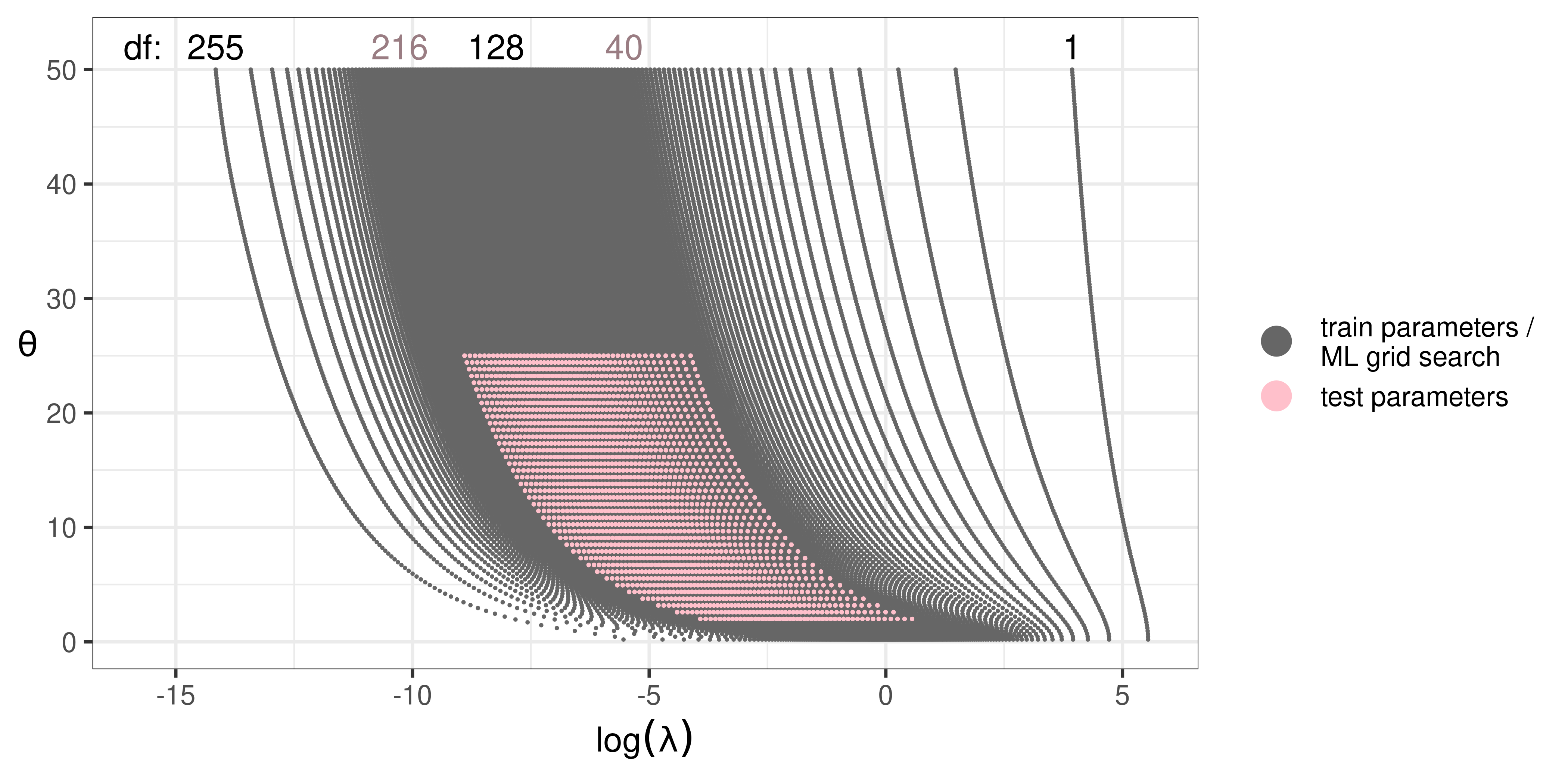}
    \caption{Scatter plot of the $\theta$ ($y$-axis) and $\log(\lambda)$ ($x$-axis) parameters.
      The $\numprint{40200}$ gray points represent the parameter configurations used to train the NNs and find the ML estimates.
      The $\numprint{2000}$ pink points represent parameter configurations used for testing.
      Each line of points corresponds to models having the same degrees of freedom (indicated on top). }
    \label{fig:grid}
\end{figure}

To detail the grid search approach to maximize~\eqref{eq:lik} we make further assumptions motivated by the data application of Section~\ref{sec:4}.
First, we set the covariance function $c(s, s^\prime, \theta)$ to be Mat\'ern with range~$\theta$ and a fixed smoothness parameter~$\nu=1$, which is a standard model used in spatial statistics.
Second, we exploit the fact that the ML estimates for $\sigma^2$ and $\tau^2$ have a closed-form given the estimate for $ \lambda =\tau^2 / \sigma^2$.
Without loss of generality, we can set $\sigma^2=1$ implying $\tau^2 = \lambda$, and hence, the parameters to be inferred are $\bxi=(\theta, \lambda)\T$.
And third, the field $\y$ consists of a $16\times 16$ fields with $n=256$ observations.
With these assumptions and the considerations on the EDFs, we construct an efficient parameter grid to maximize~\eqref{eq:lik} based on a stratified design.
More specifically, we choose a grid of $201 \times 200 = \numprint{40200}$ parameter configurations, where the $\theta$~values are equally spaced between $2$ and~$50$, and for each choice of $\theta$, $\log(\lambda)$ values are chosen such that the EDFs of the associated models are equally spaced between $1$ and~$255$; see Figure~\ref{fig:grid} for visualization. 
This design makes the $\lambda$ values comparable across different range parameters.
To find the ML estimates we evaluate the log-likelihoods for all those parameter configurations, and the configuration leading to the largest log-likelihood identifies the estimates.
Note that this design provides a robust estimate while entailing attractive computational features, such as the ability to reuse Cholesky factors among parameter configurations with the same values of $\theta$ and parallel computing opportunities.
Other numerical optimization methods might find the global optimum with fewer evaluations of the likelihood but at the price of losing some of these advantages.

\subsection{NN framework for parameter estimation}
The NN framework allows us to take a different view on this estimation problem by determining the mapping between data and parameters directly.
Instead of deriving $\cF_\text{NN}$ from~\eqref{eq:mod}, it is defined as
\begin{equation}\label{eq:composition}
   {\cal F}_\text{NN}: \IR^n \rightarrow D,\quad  {\cal F}_\text{NN}(\y) = \widehat \bxi,
\end{equation}
where $\y$ is the input data, which is in our case a sample of a $16 \times 16$ field and $\widehat\bxi$ is the corresponding estimate.
In contrast to $\cF_\text{ML}$, $\cF_\text{NN}$~depends on $p$~weights $\w=(w_1, \dots, w_p)\T$ to be determined.
With this formulation, we can express the main results of this work.
We show that ${\cal F}_\text{NN}$ can give an accurate approximation to ${\cal F}_\text{ML}$ and that ${\cal F}_\text{NN}$ can be evaluated at least two orders of magnitude faster than~$\cF_\text{ML}$.
Inferring $\w$ is also called \emph{training} the model and requires \emph{training data} consisting of $(\y, \bxi)$ pairs and a loss function~$L(\widehat\bxi,\bxi)$ that quantifies the accuracy of~$\widehat\bxi$ compared to~$\bxi$.
The weights are found by minimizing this loss function, which is a non-trivial task.
Fortunately, there are numerical optimizers built around batch gradient descent methods that perform well at this task, in part because they harness state-of-the-art CPU and GPU computing \citep{rude:17}.

The specific form of ${\cal F}_\text{NN}$ is also termed a \emph{model architecture} and is defined through the \emph{sequence} (composition) of \emph{layers} (functions) $f_1, \dots, f_l$, \ie, ${\cal F}_\text{NN} (\x) =  (f_l  \circ \cdots \circ f_1) (\x)$.
In our case, we use a combination of \emph{dense} and \emph{convolutional} layers together with \emph{linear} and \emph{ReLU} activation functions.
A dense layer with $n_I$ inputs produces $n_O$ outputs and is fully connected, \ie, it involves $\cO(n_I * n_O)$ weights.
A convolutional layer consists of $m$ linear filters of a given \emph{kernel size}, \eg, $M \times M$, which are convolved with the input field.
In this setup, there are $\cO( m* M^2) $ weights to optimize.
We refer to \cite{Hugo:etal:20} for more details on convolutional layers and \url{www.tensorflow.org/api_docs} for information on the TensorFlow implementation of both dense and convolutional layers.

\subsubsection{Training the NNs}
Training ${\cal F}_\text{NN}$ adapts this arbitrary function into one performing well at the desired task.
It involves generating training data, defining a loss function, and configuring the optimizer.  
We found that the design of the training samples is crucial for the performance of~${\cal F}_\text{NN}$, \ie,
the training samples should correspond to identifiable parameter configurations covering the parameter region of interest.
To find such parameters, we again rely on the concept of EDFs introduced in Section~\ref{sec:ml} and use the same $\numprint{40200}$ parameter configurations already used for the ML grid search; see Figure~\ref{fig:grid}.
Choosing the same parameter grid for training the NNs and the ML optimization also ensures a fair comparison of the models.
Once the parameter grid is defined, \eqref{eq:mod} can be used to simulate corresponding fields for the training samples.
Note that this design of training parameters allows us to reuse intermediate compute results, such as Cholesky factors.
Together with a vectorized implementation using the Python API to Tensorflow \citep{TF} and the GPU back-end available in Google Colab\footnote{\url{https://colab.research.google.com}}, we arrive at simulating $\numprint{40200}$ $16 \times 16$ fields within a few seconds.

The fast computations allow us to simulate training datasets consisting of multiples of $\numprint{40200}$ training samples, and we refer to such a set as an \emph{epoch}.
Also, we can refresh the training fields of an epoch often, in the extreme case every time the optimizer was informed by its samples.
This is helpful to avoid overfitting certain random features in the training fields, \ie, it supports the good performance of the model on fields not included in the training data.
However, it does not protect against overfitting the chosen covariance parameter configurations, and hence, the model is likely to perform well in parameter regions covered by many training parameter configurations. 

We use the mean absolute error (MAE) loss function to assess the accuracy of the model on a set of training samples.
To allow both $\lambda$ and $\theta$ to have a similar impact on the optimization, we scale the training parameters by their mean and standard deviation before applying the loss function, \ie,
all $\lambda_i$ and $\theta_i$ of the parameter grid are transformed according to
\begin{equation}\label{eq:scaling}
\log( \Lambda_i) = (\log(\lambda_i) - \mean_k(\log(\lambda_k)))/\sd_k(\log(\lambda_k)), \quad \Theta_i = (\theta_i - \mean_k(\theta_k))/\sd_k(\theta_k),
\end{equation}
where $\log( \Lambda_i)$ and $\Theta_i$ are now the scaled parameters, and $k=1, \dots, \numprint{40200}$ is an index such the mean and standard deviation is taken with respect to all training parameters. 
Then the loss of $q$ samples is given as
\begin{equation}\label{eq:mae}
 L\left((\widehat\bxi_1, \dots, \widehat \bxi_q),(\bxi_1, \dots,  \bxi_q)\right) = \sum_{i=1}^q \left| \log (\widehat \Lambda_i) - \log (\Lambda_i) \right| + \left|\widehat \Theta_i - \Theta_i\right|, 
\end{equation}
where $\log (\widehat \Lambda_i)$ and $\widehat \Theta_i$ are the current estimates returned by~${\cal F}_\text{NN}$.
We use this loss to inform the Adam optimizer \citep{king:17}, which updates initially random weights at a learning rate of~$0.001$.
The net result is that we can train ${\cal F}_\text{NN}$ into a useful function for the covariance parameter estimation.

\subsubsection{NN architecture featuring convolutional layers}\label{sec:ni}
Next, we give a detailed specification of a convolutional NN $\cF_\text{NN}$ taking $\y$ as input and mapping it to two scalar values, which become estimates of $\log(\lambda)$ and~$\theta$ after the training.
We call it the \emph{NF} model, which is short for \emph{NN processing fields}.
To describe its architecture, we report a sequence of layers transforming \emph{tensors} (arrays) into other tensors.
The layers have weights that determine their transformation, and it is common to report how layers transform the \emph{shape} (dimension) of the input tensors.
For example, the model input is a tensor of shape \mbox{[--, 16, 16, 1]}, where '--' stands for an arbitrary number of samples, the '16's indicate the width and height of the fields, and '1' specifies that there is one replicate.
The first convolutional layer of the NF model transforms that input tensor into another one of \emph{output} shape \mbox{[--, 7, 7, 128]} by using $128$ filters with a kernel size of $10\times 10$.
Using this notation, we describe the architecture of the NF model in Table~\ref{tab:ni}. 

\begin{table}[h]
 \caption{Summary of the NF model. It is a sequential NN taking $\y$ of shape \mbox{[--, 16, 16, 1]} as input and mapping it to two scalar values of shape~\mbox{[--, 2]}.
    After the training, the outputs become estimates of $\log(\lambda)$ and~$\theta$.}
  \label{tab:ni}
      \centering
    \begin{tabular}{lrrrlr}
        \toprule
layer type           &                  output shape    &    filters &  kernel size    &   activation&          parameters   \\    
\midrule
2D convolution        &                [--, 7, 7, 128]   &      128  &  10$\times$10 & ReLU &          \numprint{12928}        \\ 
2D convolution     &                   [--, 3, 3, 128]   &      128  &   5$\times$5&  ReLU &    \numprint{409728}        \\
2D convolution     &                   [--, 1, 1, 128]   &      128  &   3$\times$3&  ReLU &  \numprint{147584}        \\
flatten      &                         [--, 128]         &           &    &  & \numprint{0}            \\ 
dense           &                      [--, 500]         &           &    &  ReLU &    \numprint{64500}         \\
dense       &                          [--, 2]           &           &    &  linear&\numprint{1002}          \\
\midrule
\multicolumn{2}{l}{total trainable parameters:}&&&& $\numprint{635742}$\\
\bottomrule
    \end{tabular}
\end{table}

Because the training fields for the NF model are simulated using a stationary and isotropic covariance model, we can generate additional fields by flips and rotation of multiples of $90$ degrees.
Hence, we arrive at an eightfold larger (\emph{augmented}) training set at a low computational cost.
This allows us to generate epochs of \numprint{964800} training samples and to replace the fields with newly simulated ones at the end of each epoch.
We let the optimizer update the weights of the model every $200$ samples, \ie, we use a \emph{batch size} of~$200$.
With this configuration, one epoch of training takes $22$s (seconds), and training the NN for $\numprint{1000}$ epochs takes $6$h (hours) in total.
The Jupyter Notebook used to train the NN is available in the supplementary material \footnote{\url{https://github.com/florafauna/TFspatstat_paper_supplementary_material}}.
The notebook also plots the MAE over the course of the training and indicates that after $300$ epochs the optimization has (almost) converged, \ie,
the MAE decreases at a small rate after $300$~epochs.
Thus, one could reduce the training time from $6$h to $2$h without loss of accuracy in~$\cF_\text{NN}$.

For the situation where the samples consist of $30$ replicate fields $\{\y_1, \dots, \y_{30}\}$, we apply the NF model to each field, which leads to the estimates $\log(\widehat\lambda_1), \dots, \log(\widehat\lambda_{30}) $ and $\widehat\theta_1, \dots, \widehat\theta_{30}$.
Then, we take means to arrive at the estimates $\log(\widehat\lambda) = \mean_{k=1}^{30}\log(\widehat\lambda_k)$ and $\widehat\theta = exp\left(\mean_{k=1}^{30}\log(\widehat\theta_k)\right)$.
We denote this model with \emph{NF30}, which is short for \emph{NN processing $30$ replicated fields}.

\subsubsection{NN architecture featuring variograms}\label{sec:nv}
One drawback of the NF and NF30 models is that they require the input to consist of completely observed fields, \ie, the input cannot contain missing values or be observed at irregular locations.
To overcome this limitation, we present a second architecture, which first transforms $\y$ into a variogram of $119$ values before mapping it to $\log(\lambda)$ and~$\theta$.
We refer to his model as \emph{NV}, which is short for \emph{NN processing variograms}.
The network features three dense layers as detailed in Table~\ref{tab:nv}.
We choose dense over convolutional layers because dense layers capture the shape of the entire variogram.
Conversely, convolutional layers focus on the local behavior of the variogram, which encodes little information relevant to the parameter estimation.

\begin{table}[h]
  \caption{Summary of the NV and NV30 models.
    Both models are sequential NNs taking $\y$ and $\{\y_1, \dots, \y_{30}\}$ as input, respectively, and mapping it to two scalar values.
    The corresponding input shapes for the NV and NV30 models are \mbox{[--, 119]} and \mbox{[--, \numprint{3570}]}, respectively. 
    After the training, the outputs become estimates of $\log(\lambda)$ and~$\theta$.
    Although NV and NV30 models have identical layer specifications, the number of parameters of their first layer varies because of the different input shapes. 
  }
  \label{tab:nv}
      \centering
    \begin{tabular}{lrlrr}
        \toprule
layer type           &      output shape    &       activation&   parameters NV     & parameters NV30        \\    
\midrule                                                                                            
dense               &       [--, \numprint{3000}]   &  ReLU &   \numprint{360000} &\numprint{10713000}        \\
dense               &       [--, \numprint{1000}]   &  ReLU &   \numprint{3001000}&\numprint{3001000}        \\
dense               &       [--, 2]           &       linear&   \numprint{1002}   &\numprint{1002}          \\
\midrule
\multicolumn{2}{l}{total trainable parameters:}&& $\numprint{3363002}$&$\numprint{13716002}$\\
\bottomrule
    \end{tabular}
\end{table}

There are several options to adapt the NV model such that it can take $\{\y_1, \dots, \y_{30}\}$ instead of $\y$ as input.
Analog to the NF30 model, the NV model can be applied to the variograms of $\{\y_1, \dots, \y_{30}\}$ separately before taking the mean of the resulting estimates.
Another approach is to summarize the $30$ variograms of each sample by a single variogram, \eg, by taking the mean values for each distance bin.
This makes the input compatible with the NV model but ignores potentially useful information about the variability between the variograms.
In this study, we use a third approach, which overcomes this drawback by letting NV30 take all $30$ variograms as input.
Although we use the same layer specification for both NV and NV30, the larger input of $30\cdot 119=\numprint{3570}$ values implies that the first dense layer of NV30 has a considerably larger number of weights; see Table~\ref{tab:nv}.

The different number of parameters of the NV and NV30 model makes it necessary to train both models separately. 
We use epochs of $\numprint{120600}$ samples and a batch size of $200$ to train the models. 
Moreover, the spatial field to variogram transformation is computationally more expensive than simulating fields.
This is not an issue for the training of the NV model and we generate new fields and variograms at the beginning of each epoch.
However, for the NV30 model $30$ times more variogram calculations are required slowing down the training. 
Therefore, we only generate new fields and variograms every $50$th epoch, and for the other epochs simply random shuffling the samples and the variograms within a sample to generate additional variability. 
The training of the NV and NV30 models for $\numprint{1000}$ epochs takes $40$m (minutes) and $1.3$h, respectively. 
Similar to the training of the NF model, the decrease in MAE is negligible after $300$ epochs.
Thus, we could reduce the training time to $20$m and $45$m, respectively, without loss of accuracy in $\cF_\text{NN}$.
The Jupyter Notebooks used to train these models are available in the supplementary material and contain plots of the~MAEs.

\begin{figure}[!hp]
  \vspace*{.7cm}
  \centering
  \includegraphics[width=\textwidth]{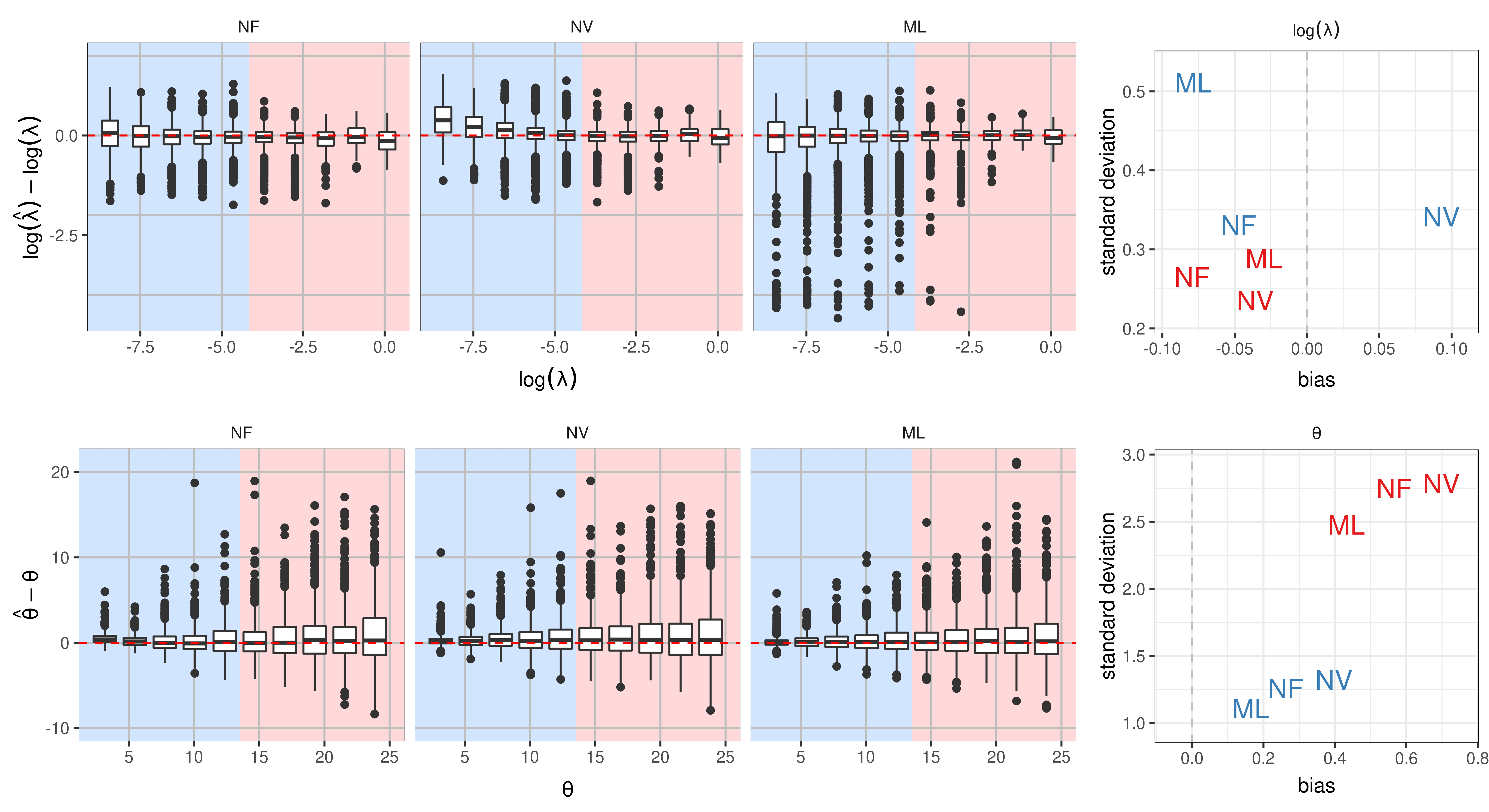}\\
  \vspace*{-10.2cm}
  \begin{flushleft}
    \hspace*{.3cm}{\large(a)}
  \end{flushleft}
  \vspace*{9.5cm}
  \includegraphics[width=\textwidth]{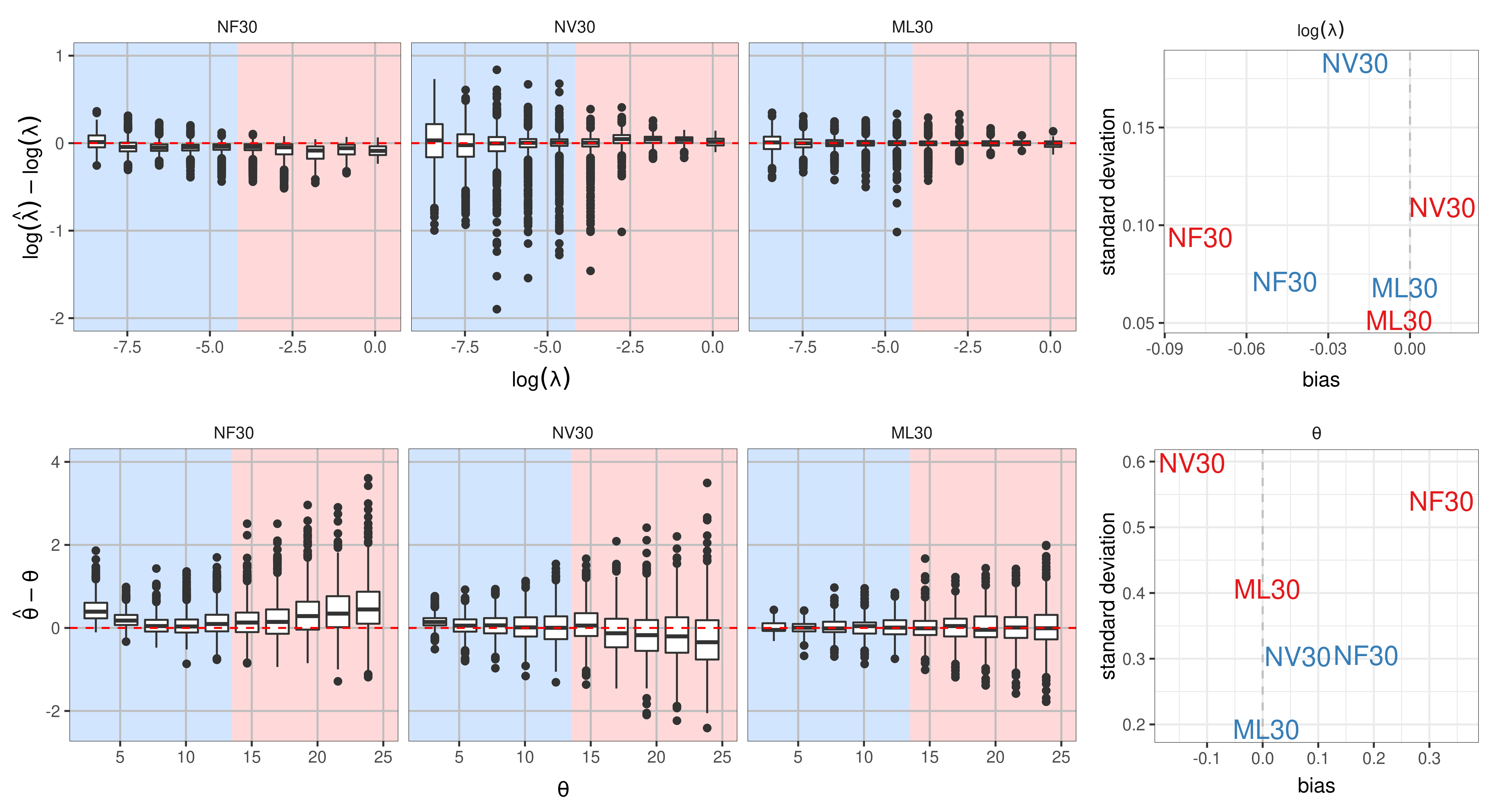}
  \vspace*{-10.5cm}
  \begin{flushleft}
    \hspace*{.2cm}{\large(b)}
  \end{flushleft}
  \vspace*{9cm}
  \caption{
    Results of the simulation study of Section~\ref{sec:3}: 
    (a) An overview of the accuracy of the $\log(\widehat\lambda)$ and $\widehat\theta$ estimates obtained by the NF, NV, and ML models.
    The box plots summarize the deviation of $\numprint{10000}$ estimates from the true values.
    The scatter plots show the bias and standard deviation of the estimates for the lower and upper half of the parameter space,
    \ie, the blue and red symbols indicate results for the lower and upper half of the parameter space, respectively.
    (b) Similar figures summarizing the estimates obtained by the NF30, NV30, and ML30 models.
  } 
  \label{fig:sim}
\end{figure}

\section{Simulation study}\label{sec:3}
Assessing the accuracy of these NNs is straight forward as one can generate a large sample of test fields. 
We use  $\numprint{2000}$ parameter configurations such that the corresponding models have a spatial range $\theta \in [2,25]$ and $\lambda$ corresponding to degrees of freedom in $[40,216]$.
For reference, the test parameters are shown as pink points in Fig.~\ref{fig:grid}.
We simulate $150$ test fields of size $16 \times 16$ for each parameter configuration using~\eqref{eq:mod}, which yields $\numprint{300000}$ test fields in total.

\subsection{Comparison of the NF, NV, and ML models}\label{sec:sim1}
The accuracy of the parameter estimates obtained by the NF, NV, and ML models is summarized in Figure~\ref{fig:sim}~a.
The box plots are based on test data with $5$ samples per test parameter configuration, \ie, $\numprint{10000}$ samples with one field in total, and indicate that the three models perform at a similar level of accuracy.
Here we use a subset of the test samples in the box plots to simplify the figures, and the impressions are confirmed by the full testing set.
We find it surprising that an empirical function can be trained to accurately reproduce these covariance parameter estimates.
A closer look, however, suggests that the estimates of $\log(\lambda)$ from the NF and NV models show some bias at the lower and upper ends of the parameter space.
This feature is less pronounced for the ML estimates, but instead, they have larger variability in the lower half of the parameter space.
The estimates of $\theta$ show little bias and a similar pattern of variability across all cases. 
Those findings are confirmed in the bias-variance scatter plots (Figure~\ref{fig:sim}~a, right panels).

Using the trained NF and NV models for parameter estimation needs much less computing resources compared to the ML model.
In particular, the low computational cost of the NF and NV models allows us to evaluate them on a laptop with four Intel Core i7-6600U CPUs @ 2.60GHz and 16~GB~RAM.
To estimate the parameters for all $\numprint{300000}$ test fields the NF and NV models take $37$s and $128$s, respectively.
Notably, the NV model spends more than half of the compute time on transforming the fields into variograms, and we suspect this part of the computation could be optimized using Fourier transformation techniques.
On the other hand, the ML model runs for $2.2$h on a server with $80$ Intel Xeon CPU E7-2850 CPUs @ 2.00GHz and 2~TB RAM to estimate all parameters.
Our implementation of the ML model is not compatible with a laptop because it relies on a large~RAM. 
However, if an implementation requiring less RAM would have similar performance, computing the estimates would take close to $2$d~(days).  
Although all presented implementations can be optimized, we believe that the timings give a realistic impression of the performance gains from using trained NN models.

\subsection{Comparison of the NF30, NV30, and ML30 models}
The accuracy of the parameter estimates obtained by the NF30, NV30, and ML30 models is summarized in Figure~\ref{fig:sim}~b.
Again the box and scatter plots are based on test data with $5$ samples per test parameter configuration, \ie, $\numprint{10000}$ test samples in total, but here each sample features $30$ independent fields. 
As expected, using $30$ independent fields reduces the variability of the estimates compared to Figure~\ref{fig:sim}~a where only one field is available.
The estimates from the NF30 and NV30 models show larger biases and variances than those from the ML30 model, especially for values of $\theta$ larger than~$15$.
But in relation to the actual size of the parameters, the biases and variances are small.
The computing resources use by the NF30, NV30, and ML30 models are similar to those of the corresponding NF, NV, and ML models given in Section~\ref{sec:sim1}.
That is, the NF30 and NV30 models achieve at least two orders of magnitude speedup over the ML30 computation.

\section{Data illustration}\label{sec:4}
\begin{figure}
  \centering
  \includegraphics[height=.3\textwidth]{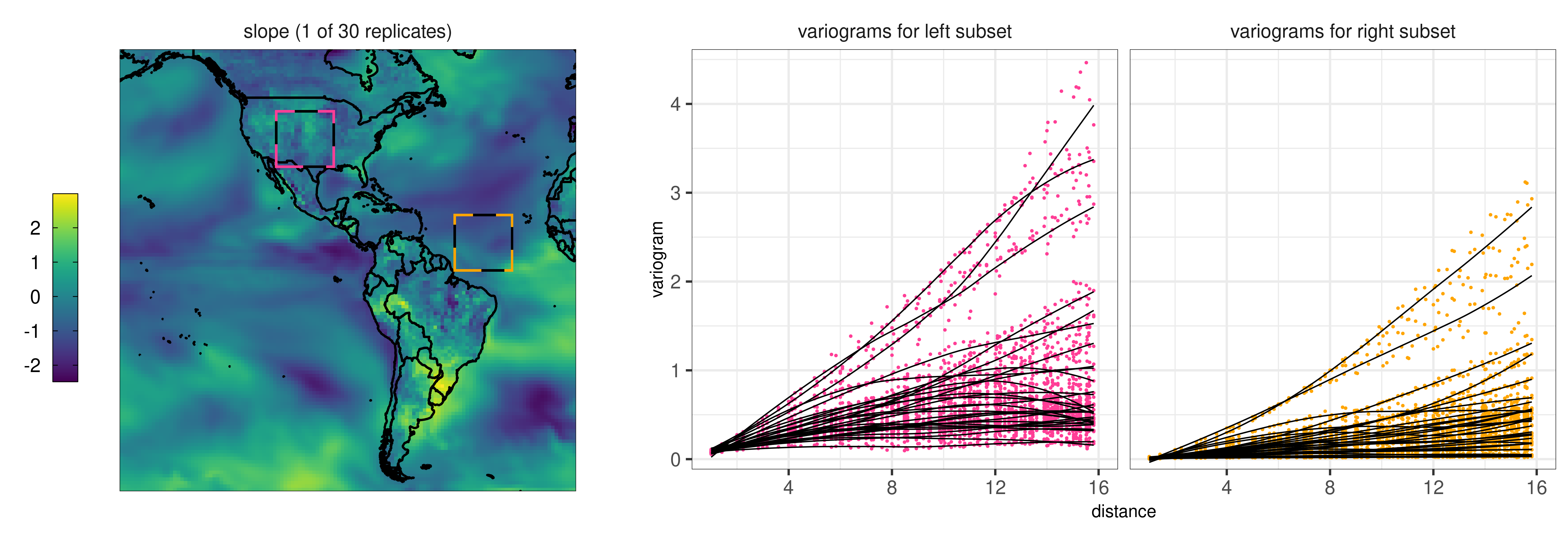}
  \vspace*{-3mm}
  \caption{Left: The scaled slope field is based on the result of one climate model run. The two squares indicate $16\times 16$ subsets.
  Right: Variograms for the left and right $16\times 16$ subset. Each of the 30 smoothing lines summarizes the variogram for one of the $30$ climate model runs.}
  \label{fig:data_eda}
\end{figure}

For the pattern scaling fields of temperature derived from the LENS 30 member ensemble, we focus on the local covariance estimates from the NV30 and ML30 models. 
The NF30 model is excluded as it shows a slightly lower performance in the simulation study compared to NV30.
The data are available from the git repository of the R~package LatticeKrig\footnote{https://github.com/NCAR/LatticeKrig/tree/master/Datasets/LENNS} and consist of $30$ spatial fields on a $288 \times 192$ regular grid (about 1.25 degrees in longitude and latitude). 
To streamline this case study, however, we consider a $128\times 128$ subset thereof, covering the North and South American continents. 
Moreover, we standardize the data by the mean and standard deviation at each location. 
For illustration, one of the $30$ standardized fields is shown in Figure~\ref{fig:data_eda} (left).
The smoother surface over the sea compared to over the land suggests a non-stationary structure. 
To take a closer look at the spatial dependency structure of all $30$ fields, we consider the $16\times 16$ regions marked by the pink and orange squares and compute variograms for both regions and all fields.
The resulting variograms are shown in Figure~\ref{fig:data_eda} (right) and indicate that the fields in the orange region tend to be smoother compared to those in the pink region.
Assuming that the $30$ replicates are independent realizations of the same data generation process and that $16\times 16$ regions thereof can be described by~\eqref{eq:mod},
we estimate the $\log(\lambda)$ and $\theta$ covariance parameters for all $\numprint{12769}$ locations where a $16 \times 16$ window is inside the region.
The spatial distribution of these estimates yields a more complete description of the non-stationarity features of the fields.
These surfaces are an end in themselves for climatological interpretation or the initial step in identifying more sophisticated non-stationary covariance functions.
In either case, it is of interest to compare the NV30 and the ML30 estimates.
 
\begin{figure}
  \includegraphics[width=\textwidth]{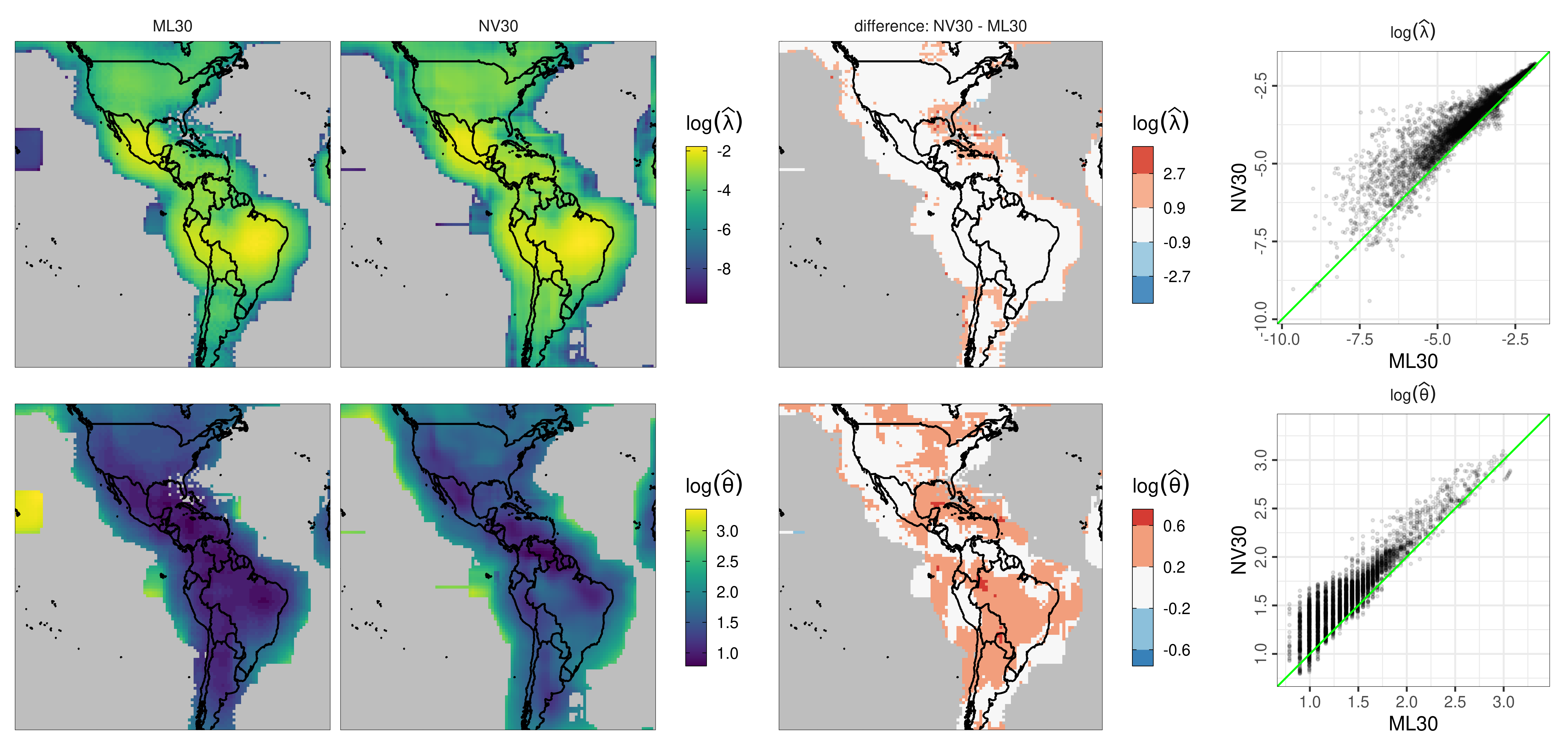}
   \vspace*{-9.3cm}
   \begin{flushleft}
    \hspace*{.1cm}(a) \hspace*{8.25cm} (b) \hspace*{5cm} (c)
   \end{flushleft}
   \vspace*{8cm}
  \caption{(a) The four panels show $\log(\widehat\lambda)$ and $\log(\widehat\theta)$ obtained from the NV30 and ML30 model.
   Gray pixels indicate parameter values with estimates outside the domain of the test parameter grid; see Fig.~\ref{fig:data_proj}. 
   (b) The difference between the NV30 and the ML30 estimates.
   Gray pixels indicate that one or both models returned estimates outside the domain of the test parameter grid.
   (c) Scatter plot of the NV30 estimates \mbox{($y$-axis)} and the ML30 estimates \mbox{($x$-axis)}. 
  }
  \label{fig:data_results}
\end{figure}

The estimates of $\log(\lambda)$ and $\log(\theta)$ are shown and compared in Figure~\ref{fig:data_results}.
The geographical comparison (Figure~\ref{fig:data_results}~a and~b) shows that the NV30 and the ML30 results produce similar estimates and geographical patterns.
However, the estimates from the NV30 model tend to be larger than those from the ML30 model.
This pattern is confirmed in the scatter plots (Figure~\ref{fig:data_results}~c), although we note that these discrepancies are amplified by the log scaling. 
While these differences can be seen as a disadvantage of the NV30 model, it has the advantage of requiring massively less computational resources, at least when the training is not counted towards the timing.
That is, based on the hardware described in Section~\ref{sec:3} estimating the parameters using NV30 takes $2$m on a laptop and using ML30 takes $2$h on a server with $80$ CPUs and 2~TB~RAM.

\begin{figure}
  \centering
 \includegraphics[width=\textwidth]{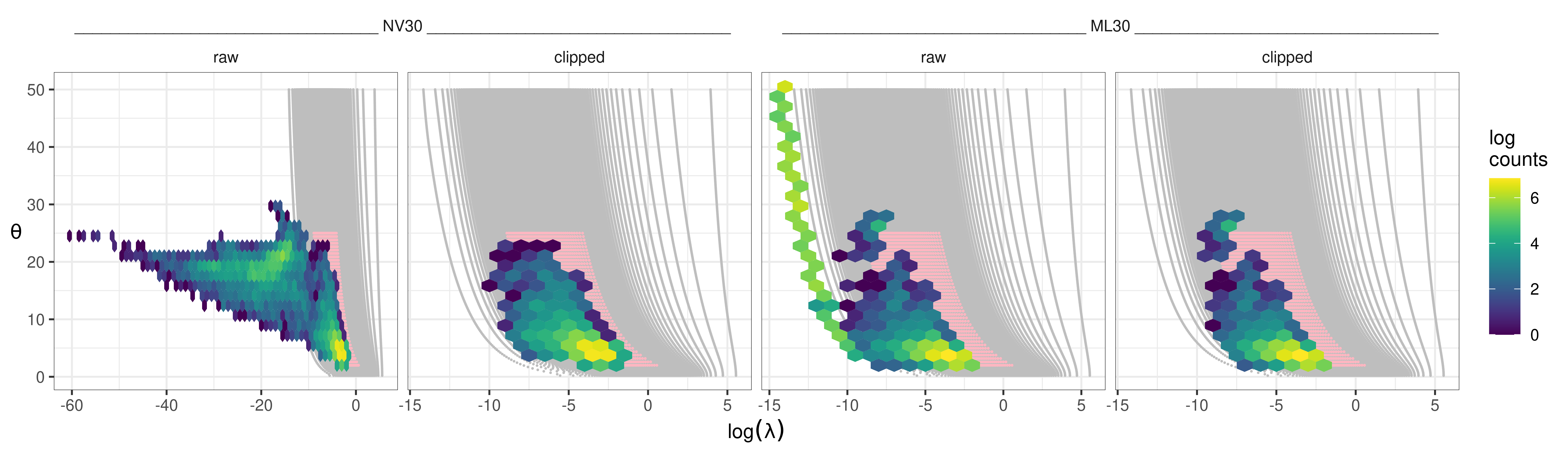}
  \vspace*{-3mm}
  \caption{Overview of the $\numprint{12769}$ estimates obtained by the NV30 and ML30 models.
   The panels indicate the number of estimates in a hexagon region of the parameter space, where the latter is spanned by $\log(\lambda)$~\mbox{($x$-axis)} and $\theta$~\mbox{($y$-axis)}.
   Gray and pink colors indicate the training parameter grid and the test parameters, respectively; compare Figure~\ref{fig:grid}.
   For both models, the raw and the clipped estimates are shown in separate panels, whereby \emph{raw} refers to the estimates returned by the model and \emph{clipped} refers to the subset thereof identified as being close to the test parameter grid.  
  }
  \label{fig:data_proj}
\end{figure}

Another way to compare these estimates is to consider the bivariate distributions of the covariance parameter estimates. 
The empirical distributions of the estimates are depicted in Figure~\ref{fig:data_proj} (panels labeled \emph{raw}).
The figure shows that a considerable fraction of the $\log(\lambda)$ estimates are tiny and correspond to models with $255$ or more out of $256$ possible degrees of freedom.
This is not surprising because one expects the pattern scaling fields to be smooth and without a white noise component over the ocean. 
While for such regions a small $\log(\widehat\lambda)$ or even $\widehat\lambda=0$ might be an appropriate estimate,
both the training parameter grid for NV30 and the grid search of ML30 are not designed to identify tiny $\lambda$ parameters.
In fact, they are optimized for parameter configurations close to the test parameters defined in Section~\ref{sec:3}.
Therefore, we do not compare the size of estimates with $\lambda$ values clearly below the test parameter configurations.
This threshold is purposeful because it is not important to distinguish among tiny choices for $\lambda$ in this application.
The remaining subsets of the estimates are shown in Figure~\ref{fig:data_proj} (panels labeled \emph{clipped}), and the geographical comparison of Figure~\ref{fig:data_results} is based on these estimates. 

\section{Discussion}\label{sec:5}
This work demonstrates that a deliberately trained NN can estimate covariance parameters from Gaussian spatial fields at an accuracy comparable to ML estimation but much faster.
In particular, when used for local estimation in a climate science application, we see speedups in computation larger than a factor of~$100$.
Although there is a limited theoretical understanding of NNs, this use can be thoroughly validated. 
The modeler has full control over the design of the training and test samples, which can be matched with the parameter region of interest, and in our case, cover a large range of parametric models.  
That stands in contrast to other applications of NNs, such as image classification, where it may be difficult to encompass the space of all possible test images.
Despite this control, there is the hazard that NN based estimates are unreliable when the model is applied to data that deviates from the statistical model used for training.
However, a careful design of the test data can help identify such situations, and the fast training and field simulation methods enable improving the model quickly.
Finally, we note that even classical ML estimates are only meaningful under the correct statistical model without further investigating the effect of model misspecification.

In this work, we have limited the modeling possibilities and followed a proof-of-concept approach.
We acknowledge that we have avoided the more classic spatial problem of estimating a global, but small, set of covariance parameters for a massive number of spatial locations.
However, this omission is due to the practical observation that huge spatial data volumes are often heterogeneous and need to be modeled locally. 
This local approach is consistent with our focused example and is useful for a substantial statistical application featuring a large spatial dataset from climate science.
Clearly, future work should entertain more complex covariance models and different configurations of spatial locations, and we believe our results strongly suggest that using NN approximations for covariance parameter estimation will be successful.
The effectiveness of using variograms as input is particularly promising in that it can address unequally spaced observations and at the same time reduces the dimension of the input to the~NN.
In this work, we compute variograms from fields over a regular grid, and the corresponding NV and NV30 models spend a considerable amount of computing resources on this calculation.
We suspect that using Fourier transformation techniques could make this computation more efficient.
Also missing from this case study are examples of including a fixed, regression part in the spatial model, but this aspect can be handled, \eg, using a back-fitting approach \citep{gerb:nych:20}. 

Although TensorFlow greatly simplifies designing and training NNs, many modeling choices can only be justified by testing them.
That leaves the modeler with a potentially time-consuming trial-and-error strategy, but we share some of our experiences:
First, the design of the training samples, and especially, the choice of the $\log(\lambda)$ and $\theta$ training parameter grids are key to a good model.
We find that taking into account the EDFs of the associated models when choosing the training parameters substantially improves the performance of the models (see Figure~\ref{fig:grid}).
Conversely, training models using a generic regular grid of $\log(\lambda)$ and $\theta$ training parameters leads to poor performance, supposedly because including very smooth and noisy fields confuses the training process.
Second, it is invaluable that the training fields can be simulated at a low computational cost using~\eqref{eq:mod}.
That allowed us to train the model with many different fields, and hence, overfitting random features of the training fields can be easily controlled.
Finally, all other modeling choices, such as the number of layers, the layer specification, as well as the optimizer and its tuning parameters have a relatively small impact on performance, and so, reduces the need to explore many architectures. 
Of course, the overall structure is relevant but whether a dense layer has $\numprint{1000}$ or $\numprint{3000}$ nodes seems less important.
We also note that biases in the NN estimates can be identified using test samples and simply be adjusted in a post-processing step based on more conventional curve-fitting approaches. 

Besides improving the performance of the NN estimates, we see many opportunities for future research on covariance parameter estimation using~NNs.
For example, one shortcoming of the presented NN models is that they do not provide any uncertainty quantification of the estimates. 
This issue could be addressed by including uncertainty measures into the NNs as additional output in combination with an adjusted training procedure. 
Going a step further, one could even let the NNs return the likelihood or likelihood surface, which would enable statistical testing in the likelihood framework. 
However, this approach will not be able to quantify uncertainty inherent to the NN models themselves, and the accuracy of the estimated likelihood values can only be quantified through testing. 
If more certainty and accuracy are required, it might be worth considering the NNs as a way to speedup ML estimation by providing good initial values for the optimization of the likelihood.
Finally, another promising direction is the extension of the approach to more complex covariance models, possibly capturing anisotropy and non-stationarity or to compute goodness-of-fit and other diagnostics statistics.

\section*{Acknowledgments}

\subsection*{Author contributions}
Florian Gerber wrote the manuscript, designed the numerical experiments, and wrote the code appearing in the supplementary material. 
Douglas Nychka provided an initial example for the NF model and collaborated on drafts of the manuscript. 

\subsection*{Financial disclosure}
Florian Gerber was supported by the Swiss National Science Foundation (grant P400P2\_186680).
Some computations were performed on the free Google Colab service (\url{https://colab.research.google.com}) and on a server at the University of Zurich, Switzerland.
The authors acknowledge C.~Tebaldi and S.~Alexeeff for their insight in developing the climate pattern scaling application.

\subsection*{Conflict of interest}

The authors declare no potential conflict of interests.

\bibliography{Main_Document}%

\begin{thebibliography}{}

\bibitem [\protect \citeauthoryear {%
Abadi%
\ \protect \BOthers {.}}{%
Abadi%
\ \protect \BOthers {.}}{%
{\protect \APACyear {2016}}%
}]{%
TF}
\APACinsertmetastar {%
TF}%
\begin{APACrefauthors}%
Abadi, M.%
, Barham, P.%
, Chen, J.%
, Chen, Z.%
, Davis, A.%
, Dean, J.%
\BDBL {}Zheng, X.%
\end{APACrefauthors}%
\unskip\
\newblock
\APACrefYearMonthDay{2016}{}{}.
\newblock
{\BBOQ}\APACrefatitle {TensorFlow: A System for Large-Scale Machine Learning}
  {Tensorflow: A system for large-scale machine learning}.{\BBCQ}
\newblock
\BIn{} \APACrefbtitle {12th {USENIX} Symposium on Operating Systems Design and
  Implementation ({OSDI} 16)} {12th {USENIX} symposium on operating systems
  design and implementation ({OSDI} 16)}\ (\BPGS\ 265--283).
\newblock
\APACaddressPublisher{}{{USENIX} Association}.
\newblock
\begin{APACrefURL}
  \url{https://www.usenix.org/conference/osdi16/technical-sessions/presentation/abadi}
  \end{APACrefURL}
\PrintBackRefs{\CurrentBib}

\bibitem [\protect \citeauthoryear {%
Alexeeff%
, Nychka%
, Sain%
\BCBL {}\ \BBA {} Tebaldi%
}{%
Alexeeff%
\ \protect \BOthers {.}}{%
{\protect \APACyear {2018}}%
}]{%
Alex:18}
\APACinsertmetastar {%
Alex:18}%
\begin{APACrefauthors}%
Alexeeff, S\BPBI E.%
, Nychka, D.%
, Sain, S\BPBI R.%
\BCBL {}\ \BBA {} Tebaldi, C.%
\end{APACrefauthors}%
\unskip\
\newblock
\APACrefYearMonthDay{2018}{Feb}{}.
\newblock
{\BBOQ}\APACrefatitle {Emulating mean patterns and variability of temperature
  across and within scenarios in anthropogenic climate change experiments}
  {Emulating mean patterns and variability of temperature across and within
  scenarios in anthropogenic climate change experiments}.{\BBCQ}
\newblock
\APACjournalVolNumPages{Climatic Change}{146}{3}{319--333}.
\newblock
\begin{APACrefDOI} 10.1007/s10584-016-1809-8 \end{APACrefDOI}
\PrintBackRefs{\CurrentBib}

\bibitem [\protect \citeauthoryear {%
Cremanns%
\ \BBA {} Roos%
}{%
Cremanns%
\ \BBA {} Roos%
}{%
{\protect \APACyear {2017}}%
}]{%
crem:etal:17}
\APACinsertmetastar {%
crem:etal:17}%
\begin{APACrefauthors}%
Cremanns, K.%
\BCBT {}\ \BBA {} Roos, D.%
\end{APACrefauthors}%
\unskip\
\newblock
\APACrefYearMonthDay{2017}{}{}.
\newblock
\APACrefbtitle {Deep {G}aussian Covariance Network.} {Deep {G}aussian
  covariance network.}
\newblock
\begin{APACrefURL} \url{https://arxiv.org/abs/1710.06202} \end{APACrefURL}
\PrintBackRefs{\CurrentBib}

\bibitem [\protect \citeauthoryear {%
Daskalakis%
, Dellaportas%
\BCBL {}\ \BBA {} Panos%
}{%
Daskalakis%
\ \protect \BOthers {.}}{%
{\protect \APACyear {2020}}%
}]{%
dask:20}
\APACinsertmetastar {%
dask:20}%
\begin{APACrefauthors}%
Daskalakis, C.%
, Dellaportas, P.%
\BCBL {}\ \BBA {} Panos, A.%
\end{APACrefauthors}%
\unskip\
\newblock
\APACrefYearMonthDay{2020}{}{}.
\newblock
\APACrefbtitle {Scalable {G}aussian Processes, with Guarantees: Kernel
  Approximations and Deep Feature Extraction.} {Scalable {G}aussian processes,
  with guarantees: Kernel approximations and deep feature extraction.}
\newblock
\begin{APACrefURL} \url{https://arxiv.org/abs/2004.01584} \end{APACrefURL}
\PrintBackRefs{\CurrentBib}

\bibitem [\protect \citeauthoryear {%
Gerber%
, de Jong%
, Schaepman%
, Schaepman-Strub%
\BCBL {}\ \BBA {} Furrer%
}{%
Gerber%
\ \protect \BOthers {.}}{%
{\protect \APACyear {2018}}%
}]{%
Gerb:Jong:Scha:Scha:Furr:18}
\APACinsertmetastar {%
Gerb:Jong:Scha:Scha:Furr:18}%
\begin{APACrefauthors}%
Gerber, F.%
, de Jong, R.%
, Schaepman, M\BPBI E.%
, Schaepman-Strub, G.%
\BCBL {}\ \BBA {} Furrer, R.%
\end{APACrefauthors}%
\unskip\
\newblock
\APACrefYearMonthDay{2018}{}{}.
\newblock
{\BBOQ}\APACrefatitle {Predicting Missing Values in Spatio-Temporal Remote
  Sensing Data} {Predicting missing values in spatio-temporal remote sensing
  data}.{\BBCQ}
\newblock
\APACjournalVolNumPages{IEEE Trans. Geosci. Remote Sens.}{}{}{}.
\newblock
\begin{APACrefDOI} 10.1109/TGRS.2017.2785240 \end{APACrefDOI}
\PrintBackRefs{\CurrentBib}

\bibitem [\protect \citeauthoryear {%
Gerber%
\ \BBA {} Nychka%
}{%
Gerber%
\ \BBA {} Nychka%
}{%
{\protect \APACyear {2021}}%
}]{%
gerb:nych:20}
\APACinsertmetastar {%
gerb:nych:20}%
\begin{APACrefauthors}%
Gerber, F.%
\BCBT {}\ \BBA {} Nychka, D\BPBI W.%
\end{APACrefauthors}%
\unskip\
\newblock
\APACrefYearMonthDay{2021}{}{}.
\newblock
{\BBOQ}\APACrefatitle {Parallel cross-validation: A scalable fitting method for
  {G}aussian process models} {Parallel cross-validation: A scalable fitting
  method for {G}aussian process models}.{\BBCQ}
\newblock
\APACjournalVolNumPages{Computational Statistics \& Data
  Analysis}{155}{}{107--113}.
\newblock
\begin{APACrefDOI} 10.1016/j.csda.2020.107113 \end{APACrefDOI}
\PrintBackRefs{\CurrentBib}

\bibitem [\protect \citeauthoryear {%
Heaton%
\ \protect \BOthers {.}}{%
Heaton%
\ \protect \BOthers {.}}{%
{\protect \APACyear {2018}}%
}]{%
Heat:Gerb:etal:18}
\APACinsertmetastar {%
Heat:Gerb:etal:18}%
\begin{APACrefauthors}%
Heaton, M\BPBI J.%
, Datta, A.%
, Finley, A\BPBI O.%
, Furrer, R.%
, Guinness, J.%
, Guhaniyogi, R.%
\BDBL {}Zammit-Mangion, A.%
\end{APACrefauthors}%
\unskip\
\newblock
\APACrefYearMonthDay{2018}{}{}.
\newblock
{\BBOQ}\APACrefatitle {A Case Study Competition Among Methods for Analyzing
  Large Spatial Data} {A case study competition among methods for analyzing
  large spatial data}.{\BBCQ}
\newblock
\APACjournalVolNumPages{JABES}{}{}{}.
\newblock
\begin{APACrefDOI} 10.1007/s13253-018-00348-w \end{APACrefDOI}
\PrintBackRefs{\CurrentBib}

\bibitem [\protect \citeauthoryear {%
Kay%
\ \protect \BOthers {.}}{%
Kay%
\ \protect \BOthers {.}}{%
{\protect \APACyear {2015}}%
}]{%
Kay:etal:15}
\APACinsertmetastar {%
Kay:etal:15}%
\begin{APACrefauthors}%
Kay, J\BPBI E.%
, Deser, C.%
, Phillips, A.%
, Mai, A.%
, Hannay, C.%
, Strand, G.%
\BDBL {}Vertenstein, M.%
\end{APACrefauthors}%
\unskip\
\newblock
\APACrefYearMonthDay{2015}{09}{}.
\newblock
{\BBOQ}\APACrefatitle {The {C}ommunity {E}arth {S}ystem {M}odel {(CESM)}
  {L}arge {E}nsemble {P}roject: A Community Resource for Studying Climate
  Change in the Presence of Internal Climate Variability} {The {C}ommunity
  {E}arth {S}ystem {M}odel {(CESM)} {L}arge {E}nsemble {P}roject: A community
  resource for studying climate change in the presence of internal climate
  variability}.{\BBCQ}
\newblock
\APACjournalVolNumPages{Bulletin of the American Meteorological
  Society}{96}{8}{1333--1349}.
\newblock
\begin{APACrefDOI} 10.1175/BAMS-D-13-00255.1 \end{APACrefDOI}
\PrintBackRefs{\CurrentBib}

\bibitem [\protect \citeauthoryear {%
Kingma%
\ \BBA {} Ba%
}{%
Kingma%
\ \BBA {} Ba%
}{%
{\protect \APACyear {2017}}%
}]{%
king:17}
\APACinsertmetastar {%
king:17}%
\begin{APACrefauthors}%
Kingma, D\BPBI P.%
\BCBT {}\ \BBA {} Ba, J.%
\end{APACrefauthors}%
\unskip\
\newblock
\APACrefYearMonthDay{2017}{}{}.
\newblock
\APACrefbtitle {Adam: A Method for Stochastic Optimization.} {Adam: A method
  for stochastic optimization.}
\newblock
\begin{APACrefURL} \url{https://arxiv.org/abs/1412.6980} \end{APACrefURL}
\PrintBackRefs{\CurrentBib}

\bibitem [\protect \citeauthoryear {%
Krizhevsky%
, Sutskever%
\BCBL {}\ \BBA {} Hinton%
}{%
Krizhevsky%
\ \protect \BOthers {.}}{%
{\protect \APACyear {2017}}%
}]{%
Kriz:etal:17}
\APACinsertmetastar {%
Kriz:etal:17}%
\begin{APACrefauthors}%
Krizhevsky, A.%
, Sutskever, I.%
\BCBL {}\ \BBA {} Hinton, G\BPBI E.%
\end{APACrefauthors}%
\unskip\
\newblock
\APACrefYearMonthDay{2017}{{\APACmonth{05}}}{}.
\newblock
{\BBOQ}\APACrefatitle {ImageNet Classification with Deep Convolutional Neural
  Networks} {Imagenet classification with deep convolutional neural
  networks}.{\BBCQ}
\newblock
\APACjournalVolNumPages{Commun. ACM}{60}{6}{84--90}.
\newblock
\begin{APACrefDOI} 10.1145/3065386 \end{APACrefDOI}
\PrintBackRefs{\CurrentBib}

\bibitem [\protect \citeauthoryear {%
H.~{Liu}%
, {Ong}%
, {Shen}%
\BCBL {}\ \BBA {} {Cai}%
}{%
H.~{Liu}%
\ \protect \BOthers {.}}{%
{\protect \APACyear {2020}}%
}]{%
Liu:etal:20}
\APACinsertmetastar {%
Liu:etal:20}%
\begin{APACrefauthors}%
{Liu}, H.%
, {Ong}, Y\BPBI S.%
, {Shen}, X.%
\BCBL {}\ \BBA {} {Cai}, J.%
\end{APACrefauthors}%
\unskip\
\newblock
\APACrefYearMonthDay{2020}{}{}.
\newblock
{\BBOQ}\APACrefatitle {When {G}aussian Process Meets Big Data: A Review of
  Scalable {GPs}} {When {G}aussian process meets big data: A review of scalable
  {GPs}}.{\BBCQ}
\newblock
\APACjournalVolNumPages{IEEE Transactions on Neural Networks and Learning
  Systems}{31}{11}{4405--4423}.
\newblock
\begin{APACrefDOI} 10.1109/TNNLS.2019.2957109 \end{APACrefDOI}
\PrintBackRefs{\CurrentBib}

\bibitem [\protect \citeauthoryear {%
K.~{Liu}%
, {Ok}%
, {Vega-Brown}%
\BCBL {}\ \BBA {} {Roy}%
}{%
K.~{Liu}%
\ \protect \BOthers {.}}{%
{\protect \APACyear {2018}}%
}]{%
Liu:etal:18}
\APACinsertmetastar {%
Liu:etal:18}%
\begin{APACrefauthors}%
{Liu}, K.%
, {Ok}, K.%
, {Vega-Brown}, W.%
\BCBL {}\ \BBA {} {Roy}, N.%
\end{APACrefauthors}%
\unskip\
\newblock
\APACrefYearMonthDay{2018}{}{}.
\newblock
{\BBOQ}\APACrefatitle {Deep Inference for Covariance Estimation: Learning
  {G}aussian Noise Models for State Estimation} {Deep inference for covariance
  estimation: Learning {G}aussian noise models for state estimation}.{\BBCQ}
\newblock
\BIn{} \APACrefbtitle {{2018 IEEE International Conference on Robotics and
  Automation (ICRA)}} {{2018 IEEE International Conference on Robotics and
  Automation (ICRA)}}\ (\BPGS\ 1436--1443).
\newblock
\begin{APACrefDOI} 10.1109/ICRA.2018.8461047 \end{APACrefDOI}
\PrintBackRefs{\CurrentBib}

\bibitem [\protect \citeauthoryear {%
{Lopez Pinaya}%
, Vieira%
, Garcia-Dias%
\BCBL {}\ \BBA {} Mechelli%
}{%
{Lopez Pinaya}%
\ \protect \BOthers {.}}{%
{\protect \APACyear {2020}}%
}]{%
Hugo:etal:20}
\APACinsertmetastar {%
Hugo:etal:20}%
\begin{APACrefauthors}%
{Lopez Pinaya}, W\BPBI H.%
, Vieira, S.%
, Garcia-Dias, R.%
\BCBL {}\ \BBA {} Mechelli, A.%
\end{APACrefauthors}%
\unskip\
\newblock
\APACrefYearMonthDay{2020}{}{}.
\newblock
{\BBOQ}\APACrefatitle {Chapter 10 - Convolutional neural networks} {Chapter 10
  - convolutional neural networks}.{\BBCQ}
\newblock
\BIn{} A.~Mechelli\ \BBA {} S.~Vieira\ (\BEDS), \APACrefbtitle {Machine
  Learning} {Machine learning}\ (\BPGS\ 173--191).
\newblock
\APACaddressPublisher{}{Academic Press}.
\newblock
\begin{APACrefDOI} 10.1016/B978-0-12-815739-8.00010-9 \end{APACrefDOI}
\PrintBackRefs{\CurrentBib}

\bibitem [\protect \citeauthoryear {%
Nychka%
, Hammerling%
, Krock%
\BCBL {}\ \BBA {} Wiens%
}{%
Nychka%
\ \protect \BOthers {.}}{%
{\protect \APACyear {2018}}%
}]{%
Nych:etal:18}
\APACinsertmetastar {%
Nych:etal:18}%
\begin{APACrefauthors}%
Nychka, D.%
, Hammerling, D.%
, Krock, M.%
\BCBL {}\ \BBA {} Wiens, A.%
\end{APACrefauthors}%
\unskip\
\newblock
\APACrefYearMonthDay{2018}{}{}.
\newblock
{\BBOQ}\APACrefatitle {Modeling and emulation of nonstationary {G}aussian
  fields} {Modeling and emulation of nonstationary {G}aussian fields}.{\BBCQ}
\newblock
\APACjournalVolNumPages{Spatial Statistics}{28}{}{21--38}.
\newblock
\begin{APACrefDOI} 10.1016/j.spasta.2018.08.006 \end{APACrefDOI}
\PrintBackRefs{\CurrentBib}

\bibitem [\protect \citeauthoryear {%
Ojha%
, Abraham%
\BCBL {}\ \BBA {} Snášel%
}{%
Ojha%
\ \protect \BOthers {.}}{%
{\protect \APACyear {2017}}%
}]{%
Ojha:etal:17}
\APACinsertmetastar {%
Ojha:etal:17}%
\begin{APACrefauthors}%
Ojha, V\BPBI K.%
, Abraham, A.%
\BCBL {}\ \BBA {} Snášel, V.%
\end{APACrefauthors}%
\unskip\
\newblock
\APACrefYearMonthDay{2017}{}{}.
\newblock
{\BBOQ}\APACrefatitle {Metaheuristic design of feedforward neural networks: A
  review of two decades of research} {Metaheuristic design of feedforward
  neural networks: A review of two decades of research}.{\BBCQ}
\newblock
\APACjournalVolNumPages{Engineering Applications of Artificial
  Intelligence}{60}{}{97--116}.
\newblock
\begin{APACrefDOI} 10.1016/j.engappai.2017.01.013 \end{APACrefDOI}
\PrintBackRefs{\CurrentBib}

\bibitem [\protect \citeauthoryear {%
Ruder%
}{%
Ruder%
}{%
{\protect \APACyear {2017}}%
}]{%
rude:17}
\APACinsertmetastar {%
rude:17}%
\begin{APACrefauthors}%
Ruder, S.%
\end{APACrefauthors}%
\unskip\
\newblock
\APACrefYearMonthDay{2017}{}{}.
\newblock
\APACrefbtitle {An overview of gradient descent optimization algorithms.} {An
  overview of gradient descent optimization algorithms.}
\newblock
\begin{APACrefURL} \url{https://arxiv.org/abs/1609.04747} \end{APACrefURL}
\PrintBackRefs{\CurrentBib}

\bibitem [\protect \citeauthoryear {%
Schmidhuber%
}{%
Schmidhuber%
}{%
{\protect \APACyear {2015}}%
}]{%
Schm:15}
\APACinsertmetastar {%
Schm:15}%
\begin{APACrefauthors}%
Schmidhuber, J.%
\end{APACrefauthors}%
\unskip\
\newblock
\APACrefYearMonthDay{2015}{}{}.
\newblock
{\BBOQ}\APACrefatitle {Deep learning in neural networks: An overview} {Deep
  learning in neural networks: An overview}.{\BBCQ}
\newblock
\APACjournalVolNumPages{Neural Networks}{61}{}{85--117}.
\newblock
\begin{APACrefDOI} 10.1016/j.neunet.2014.09.003 \end{APACrefDOI}
\PrintBackRefs{\CurrentBib}

\bibitem [\protect \citeauthoryear {%
Williams%
}{%
Williams%
}{%
{\protect \APACyear {1996}}%
}]{%
Pete:96}
\APACinsertmetastar {%
Pete:96}%
\begin{APACrefauthors}%
Williams, P\BPBI M.%
\end{APACrefauthors}%
\unskip\
\newblock
\APACrefYearMonthDay{1996}{}{}.
\newblock
{\BBOQ}\APACrefatitle {Using Neural Networks to Model Conditional Multivariate
  Densities} {Using neural networks to model conditional multivariate
  densities}.{\BBCQ}
\newblock
\APACjournalVolNumPages{Neural Computation}{8}{4}{843--854}.
\newblock
\begin{APACrefDOI} 10.1162/neco.1996.8.4.843 \end{APACrefDOI}
\PrintBackRefs{\CurrentBib}

\bibitem [\protect \citeauthoryear {%
Yuan%
, Deng%
, Zhang%
\BCBL {}\ \BBA {} Qu%
}{%
Yuan%
\ \protect \BOthers {.}}{%
{\protect \APACyear {2020}}%
}]{%
Yuan:etal:20}
\APACinsertmetastar {%
Yuan:etal:20}%
\begin{APACrefauthors}%
Yuan, Y.%
, Deng, Y.%
, Zhang, Y.%
\BCBL {}\ \BBA {} Qu, A.%
\end{APACrefauthors}%
\unskip\
\newblock
\APACrefYearMonthDay{2020}{}{}.
\newblock
{\BBOQ}\APACrefatitle {Deep learning from a statistical perspective} {Deep
  learning from a statistical perspective}.{\BBCQ}
\newblock
\APACjournalVolNumPages{Stat}{9}{1}{e294}.
\newblock
\begin{APACrefDOI} 10.1002/sta4.294 \end{APACrefDOI}
\PrintBackRefs{\CurrentBib}

\end{thebibliography}

\section*{Author Biography}

\begin{biography}{\includegraphics[width=70pt,height=70pt]{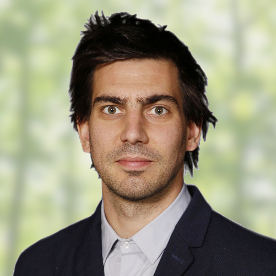}}{\textbf{Florian Gerber}
    received the BSc in Mathematics from the University of Bern, Switzerland in 2010, the MSc in Biostatistics from the University of Zurich (UZH), Switzerland in 2013, and the PhD in Applied Statistics from the Institute of Mathematics at UZH in 2017.
    He has worked on several research projects related to statistics and computing with a focus on environmental and medical applications. 
    The work presented in this manuscript was developed during his time as a PostDoc at the Colorado School of Mines, USA.
    The position was supported by PostDoc Mobility grants of the Swiss National Science Foundation. 
  }
\end{biography}

\begin{biography}{\includegraphics[width=70pt,height=86pt, clip, trim=0 1cm 0 0]{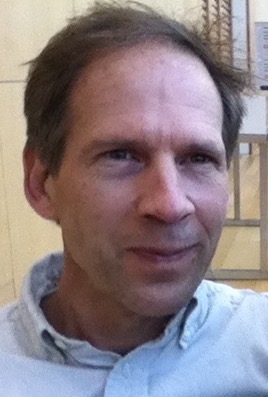}}{\textbf{Douglas W.\ Nychka}
    received his PhD in Statistics in 1983 from the University of Wisconsin.
    He has been Professor at North Carolina State University, Senior Scientist and Institute Director at the National Center for Atmospheric Research and currently is Professor in the Department of Applied Mathematics and Statistics at the Colorado School of Mines.
    He has an interest in any application of spatial statistics to the environment, curve and surface fitting problems in science and engineering, and is the primary author of the R packages \emph{fields} and \emph{LatticeKrig}.
    He is a Fellow of the American Statistical Association and the Institute of Mathematical Statistics.
}
\end{biography}

\end{document}